# Sparse Q-learning with Mirror Descent

**Sridhar Mahadevan** and **Bo Liu**
Computer Science Department
University of Massachusetts, Amherst
Amherst, Massachusetts, 01003
{mahadeva, boliu}@cs.umass.edu

## Abstract

This paper explores a new framework for reinforcement learning based on *online convex optimization*, in particular *mirror descent* and related algorithms. Mirror descent can be viewed as an enhanced gradient method, particularly suited to minimization of convex functions in high-dimensional spaces. Unlike traditional gradient methods, mirror descent undertakes gradient updates of weights in both the dual space and primal space, which are linked together using a Legendre transform. Mirror descent can be viewed as a proximal algorithm where the distance generating function used is a Bregman divergence. A new class of *proximal-gradient* based temporal-difference (TD) methods are presented based on different Bregman divergences, which are more powerful than regular TD learning. Examples of Bregman divergences that are studied include p-norm functions, and Mahalanobis distance based on the covariance of sample gradients. A new family of sparse mirror-descent reinforcement learning methods are proposed, which are able to find sparse fixed points of an $l_1$-regularized Bellman equation at significantly less computational cost than previous methods based on second-order matrix methods. An experimental study of mirror-descent reinforcement learning is presented using discrete and continuous Markov decision processes.

*Reinforcement learning* (RL) models the interaction between the agent and an environment as a *Markov decision process* (MDP), a widely adopted model of sequential decision-making. The major aim of this paper is to investigate a new framework for reinforcement learning and sequential decision-making, based on ideas from *online convex optimization*, in particular mirror descent [NY83] and related algorithms [BT03, Nes09]. Mirror descent generalizes classical first-order gradient descent to non-Euclidean geometries by using a distance-generating function specific to a particular geometry. Mirror descent can be viewed as a proximal algorithm [BT03], where the distance generating function used is a Bregman divergence [Bre67]. Mirror descent is a powerful framework for convex optimization in high-dimensional spaces: an early success of this approach was its use in positron-emission tomography (PET) imaging involving an optimization problem with millions of variables [BTMN01]. The mirror descent algorithm has led to new methods for sparse classification and regression [DHS11, SST11a]. Mirror descent has also been extended from its original deterministic setting [NY83] to a stochastic approximation setting [NJLS09], which makes it highly appropriate for RL, as the standard temporal-difference (TD) learning method for solving MDPs also has its origins in stochastic approximation theory [Bor08].

$l_1$ regularized methods for solving MDPs have become a topic of recent attention, including a technique combining least-squares TD (LSTD) with least-angle regression (LARS) [KN09]; another method for combining $l_1$ regularization with approximate linear programming [PTPZ10]; finally, a linear complementarity formulation of $l_1$ regularization [JPWP10]. These methods involve matrix inversion, requiring near cubic complexity in the number of (active) features. In contrast, mirror-descent based RL methods promise similar performance guarantees involving only linear complexity in the number of features. Recent work in online learning [DHS11, SST11a] has explored the application of mirror descent in developing sparse methods for regularized classification and regression. This paper investigates the use of mirror-descent for sparse reinforcement learning.

## 1 Technical Background

### 1.1 Reinforcement Learning

The most popular and widely used RL method is temporal-difference (TD) learning [Sut88]. TD is a stochastic approximation approach [Bor08] to solving Markov decision processes (MDPs), comprised of a set of states $S$, a set of (possibly state-dependent) actions $A$ ($A_s$), a dynamical system model comprised of the transition probabilities $P^a_{ss'}$ specifying the probability of transition to state $s'$ from state $s$ under action $a$, and a reward model $R$. A policy $\pi : S \rightarrow A$ is a deterministic mapping from states to actions. Associated with each policy $\pi$ is a value function $V^\pi$, which is a fixed point of the Bellman equation:

$$V^\pi = T^\pi(V^\pi) = R^\pi + \gamma P^\pi V^\pi \qquad (1)$$

where $0 \leq \gamma < 1$ is a discount factor. Any optimal policy $\pi^*$ defines the same unique optimal value function $V^*$ that satisfies the nonlinear system of equations:

$$V^*(s) = \max_a \sum_{s'} P^a_{ss'} (R^a_{ss'} + \gamma V^*(s'))$$

The *action value* $Q^*(s,a)$ represents a convenient reformulation of the value function, defined as the long-term value of performing $a$ first, and then acting optimally according to $V^*$:

$$Q^*(s,a) = E\left(r_{t+1} + \gamma \max_{a'} Q^*(s_{t+1}, a')|s_t = s, a_t = a\right) \qquad (2)$$

where $r_{t+1}$ is the actual reward received at the next time step, and $s_{t+1}$ is the state resulting from executing action $a$ in state $s_t$. The (optimal) action value formulation is convenient because it can be approximately solved by a temporal-difference (TD) learning technique called Q-learning [Wat89]. The simplest TD method, called TD(0), estimates the value function associated with the fixed policy using a normal stochastic gradient iteration:

$$V_{t+1}(s_t) = V_t(s_t) + \alpha_t(r_t + \gamma V_t(s_{t+1}) - V_t(s_t))$$

TD(0) converges to the optimal value function $V^\pi$ for policy $\pi$ as long as the samples are "on-policy", namely following the stochastic Markov chain associated with the policy; and the learning rate $\alpha_t$ is decayed according to the Robbins-Monro conditions in stochastic approximation theory: $\sum_t \alpha_t = \infty, \sum_t \alpha_t^2 < \infty$ [BT96]. When the set of states $S$ is large, it is often necessary to approximate the value function $V$ using a set of handcrafted basis functions (e.g., polynomials, radial basis functions, wavelets etc.) or automatically generated basis functions [Mah09]. In linear value function approximation, the value function is assumed to lie in the linear span of the basis function matrix $\Phi$ of dimension $|S| \times p$, where it is assumed that $p \ll |S|$. Hence, $V^\pi \approx \hat{V}^\pi = \Phi w$. The equivalent TD(0) rule for linear function approximated value functions is given as:

$$w_{t+1} = \alpha_t \left(r_t + \gamma \phi(s_{t+1})^T w_t - \phi(s_t)^T w_t\right) \phi(s_t) \qquad (3)$$

where the quantity in the parenthesis is the TD error. TD can be shown to converge to the fixed point of the composition of the projector $\Pi^\Phi$ onto the column space of $\Phi$ and the Bellman operator $T^\pi$.

$l_1$ regularized least-squares methods for solving MDPs attempt to find a fixed point of the $l_1$ penalized Bellman equation:[1] [KN09, PTPZ10, JPWP10]

$$w = f(w) = \mathrm{argmin}_u \left(\|(R^\pi + \gamma P^\pi \Phi w - \Phi u)\|^2 + \beta \|u\|_1\right) \qquad (4)$$

Unfortunately, the above $l_1$ regularized fixed point is not a convex optimization problem. Another way to introduce $l_1$ regularization is to penalize the projected Bellman residual [GS11], which yields a convex optimization problem:

$$\begin{aligned} w_\theta &= \mathrm{argmin}_w \|R^\pi + \gamma P^\pi \Phi \theta - \Phi w\|^2 \\ \theta^* &= \mathrm{argmin}_\theta \|\Phi w_\theta - \Phi \theta\|_2^2 + \beta \|\theta\|_1 \end{aligned} \qquad (5)$$

where the first step is the projection step and the second is the fixed point step. Note that in the on-policy setting, algorithms derived from the motivation of minimizing projected Bellman residual asymptotically converge to the solution of conventional TD-learning [SMP+09]. Recent $l_1$-regularized RL methods, such as LARS-TD and LCP-TD, involve matrix inversion, requiring at least cubic complexity in the number of (active) features. In contrast, mirror-descent based RL methods can provide similar performance guarantees involving only linear complexity in the number of features.

### 1.2 Online Convex Optimization

Online convex optimization [Zin03] explores the use of first-order gradient methods for solving convex optimization problems. In this framework, at each step, the learner picks an element $w_t$ of a convex set, in response to which an adversary picks a convex loss function $f_t(w_t)$. The aim of the learner is to select elements in such a way as to minimize its long-term regret

$$\sum_{t=1}^T f_t(w_t) - \min_w \sum_{t=1}^T f_t(w)$$

with respect to the choice it would have made in hindsight.

A fundamental problem addressed in this paper is to minimize a *sum* of two functions, as in Equation 4 and Equation 5:

$$w^* = \mathrm{argmin}_{w \in X} (f(w) + g(w)) \qquad (6)$$

where $f(w)$ is a smooth differentiable convex function, whereas $g$ is a (possibly non-smooth) convex function that is subdifferentiable over its domain. A common example is $l_1$ regularized classification or regression:

$$w^* = \mathrm{argmin}_{w \in \mathbb{R}^d} \sum_{t=1}^m L(\langle w, x_t \rangle, y_t) + \beta \|w\|_1 \qquad (7)$$

---

[1] In practice, the full $\Phi$ matrix is never constructed, and only the $p$-dimensional embedding $\phi(s)$ of sampled states are explicitly formed. Also, $R^\pi$, $\Phi$, and $P^\pi \Phi$ are approximated by $\tilde{R}$, $\tilde{\Phi}$, and $\tilde{\Phi}'$ over a given set of samples.

where $L(a, b)$ is a convex loss function, and $\beta$ is a sparsity-controlling parameter. The simplest online convex algorithm is based on the classic gradient descent procedure for minimizing a function, given as:

$$w_0 \in X, w_t = \Pi_X \left( w_{t-1} - \alpha_t \nabla f(w_{t-1}) \right) : t \geq 1 \quad (8)$$

where $\Pi_X(x) = \operatorname{argmin}_{y \in X} \|x - y\|^2$ is the projector onto set $X$, and $\alpha_t$ is a stepsize. If $f$ is not differentiable, then the subgradient $\partial f$ can be substituted instead, resulting in the well-known projected subgradient method, a workhorse of nonlinear programming [Ber99]. We discuss a general framework for minimizing Equation 6 next.

### 1.3 Proximal Mappings and Mirror Descent

The proximal mapping associated with a convex function $h$ is defined as:

$$\operatorname{prox}_h(x) = \operatorname{argmin}_u \left( h(u) + \|u - x\|_2^2 \right)$$

If $h(x) = 0$, then $\operatorname{prox}_h(x) = x$, the identity function. If $h(x) = I_C(x)$, the indicator function for a convex set $C$, then $\operatorname{prox}_{I_C}(x) = \Pi_C(x)$, the projector onto set $C$. For learning sparse representations, the case when $h(w) = \beta\|w\|_1$ is particularly important. In this case, the entry-wise proximal operator is:

$$\operatorname{prox}_h(w)_i = \begin{cases} w_i - \beta, & \text{if } w_i > \beta \\ 0, & \text{if } |w_i| \leq \beta \\ w_i + \beta, & \text{otherwise} \end{cases} \quad (9)$$

An interesting observation follows from noting that the projected subgradient method (Equation 8) can be written equivalently using the proximal mapping as:

$$w_{t+1} = \operatorname{argmin}_{w \in X} \left( \langle w, \partial f(w_t) \rangle + \frac{1}{2\alpha_t} \|w - w_t\|_2^2 \right) \quad (10)$$

An intuitive way to understand this equation is to view the first term as requiring the next iterate $w_{t+1}$ to move in the direction of the (sub) gradient of $f$ at $w_t$, whereas the second term requires that the next iterate $w_{t+1}$ not move too far away from the current iterate $w_t$. Note that the (sub)gradient descent is a special case of Equation (10) with Euclidean distance setup.

With this introduction, we can now introduce the main concept of *mirror descent* [NY83]. We follow the treatment in [BT03] in presenting the mirror descent algorithm as a nonlinear proximal method based on a distance generating function that is a Bregman divergence [Bre67].

**Definition 1:** A distance generating function $\psi(x)$ is defined as a continuously differentiable strongly convex function (with modulus $\sigma$) which satisfies:

$$\langle x' - x, \nabla \psi(x') - \nabla \psi(x) \rangle \geq \sigma \|x' - x\|^2 \quad (11)$$

Given such a function $\psi$, the Bregman divergence associated with it is defined as:

$$D_\psi(x, y) = \psi(x) - \psi(y) - \langle \nabla \psi(y), x - y \rangle \quad (12)$$

Intuitively, the Bregman divergence measures the difference between the value of a strongly convex function $\psi(x)$ and the estimate derived from the first-order Taylor series expansion at $\psi(y)$. Many widely used distance measures turn out to be special cases of Bregman divergences, such as Euclidean distance (where $\psi(x) = \frac{1}{2}\|x\|_2^2$) and Kullback Liebler divergence (where $\psi(x) = \sum_i x_i \log_2 x_i$, the positive entropy function). In general, Bregman divergences are non-symmetric, but projections onto a convex set with respect to a Bregman divergence is well-defined.

The general mirror descent procedure can be written as:

$$w_{t+1} = \operatorname{argmin}_{w \in X} \left( \langle w, \partial f(w_t) \rangle + \frac{1}{\alpha_t} D_\psi(w, w_t) \right) \quad (13)$$

Notice that the squared distance term in Equation 10 has been generalized to a Bregman divergence. The solution to this optimization problem can be stated succinctly as the following generalized gradient descent algorithm, which forms the core procedure in mirror descent:

$$w_{t+1} = \nabla \psi^* \left( \nabla \psi(w_t) - \alpha_t \partial f(w_t) \right) \quad (14)$$

Here, $\psi^*$ is the Legendre transform of the strongly convex function $\psi$, which is defined as

$$\psi^*(y) = \sup_{x \in X} \left( \langle x, y \rangle - \psi(x) \right)$$

It can be shown that $\nabla \psi^* = (\nabla \psi)^{-1}$ [BT03]. Mirror descent is a powerful first-order optimization method that been shown to be "universal" in that if a problem is online learnable, it leads to a low-regret solution using mirror descent [SST11b]. It is shown in [BTMN01] that the mirror descent procedure specified in Equation 14 with the Bregman divergence defined by the *p-norm* function [Gen03], defined below, can outperform regular projected subgradient method by a factor $\frac{n}{\log n}$ where $n$ is the dimensionality of the space. For high-dimensional spaces, this ratio can be quite large.

## 2 Proposed Framework: Mirror Descent RL

Algorithm 1 describes the proposed mirror-descent TD($\lambda$) method.[2] Unlike regular TD, the weights are updated using the TD error in the dual space by mapping the primal weights $w$ using a gradient of a strongly convex function $\psi$. Subsequently, the updated dual weights are converted

---
[2] All the algorithms described extend to the action-value case where $\phi(s)$ is replaced by $\phi(s, a)$.

**Algorithm 1** Adaptive Mirror Descent TD($\lambda$)

Let $\pi$ be some fixed policy for an MDP M, and $s_0$ be the initial state. Let $\Phi$ be some fixed or automatically generated basis.

1: **repeat**
2:  Do action $\pi(s_t)$ and observe next state $s_{t+1}$ and reward $r_t$.
3:  Update the eligibility trace $e_t \leftarrow e_t + \lambda\gamma\phi(s_t)$
4:  Update the dual weights $\theta_t$ for a linear function approximator:

$$\theta_{t+1} = \nabla\psi_t(w_t) + \alpha_t(r_t + \gamma\phi(s_{t+1})^T w_t - \phi(s_t)^T w_t)e_t$$

   where $\psi$ is a distance generating function.
5:  Set $w_{t+1} = \nabla\psi_t^*(\theta_{t+1})$ where $\psi^*$ is the Legendre transform of $\psi$.
6:  Set $t \leftarrow t + 1$.
7: **until done**.
   Return $\hat{V}^\pi \approx \Phi w_t$ as the value function associated with policy $\pi$ for MDP $M$.

back into the primal space using the gradient of the Legendre transform of $\psi$, namely $\nabla\psi^*$. Algorithm 1 specifies the mirror descent TD($\lambda$) algorithm wherein each weight $w_i$ is associated with an eligibility trace $e(i)$. For $\lambda = 0$, this is just the features of the current state $\phi(s_t)$, but for nonzero $\lambda$, this corresponds to a decayed set of features proportional to the recency of state visitations. Note that the distance generating function $\psi_t$ is a function of time.

### 2.1 Choice of Bregman Divergence

We now discuss various choices for the distance generating function in Algorithm 1. In the simplest case, suppose $\psi(w) = \frac{1}{2}\|w\|_2^2$, the Euclidean length of $w$. In this case, it is easy to see that mirror descent TD($\lambda$) corresponds to regular TD($\lambda$), since the gradients $\nabla\psi$ and $\nabla\psi^*$ correspond to the identity function. A much more interesting choice of $\psi$ is $\psi(w) = \frac{1}{2}\|w\|_q^2$, and its conjugate Legendre transform $\psi^*(w) = \frac{1}{2}\|w\|_p^2$. Here, $\|w\|_q = \left(\sum_j |w_j|^q\right)^{\frac{1}{q}}$, and $p$ and $q$ are conjugate numbers such that $\frac{1}{p} + \frac{1}{q} = 1$. This $\psi(w)$ leads to the p-norm link function $\theta = f(w)$ where $f : \mathbb{R}^d \to \mathbb{R}^d$ [Gen03]:

$$f_j(w) = \frac{\text{sign}(w_j)|w_j|^{q-1}}{\|w\|_q^{q-2}}, \quad f_j^{-1}(\theta) = \frac{\text{sign}(\theta_j)|\theta_j|^{p-1}}{\|\theta\|_p^{p-2}} \quad (15)$$

The p-norm function has been extensively studied in the literature on online learning [Gen03], and it is well-known that for large $p$, the corresponding classification or regression method behaves like a multiplicative method (e.g., the p-norm regression method for large $p$ behaves like an exponentiated gradient method (EG) [KW95, Lit88]).

Another distance generating function is the negative entropy function $\psi(w) = \sum_i w_i \log w_i$, which leads to the entropic mirror descent algorithm [BT03]. Interestingly, this special case has been previously explored [PS95] as the exponentiated-gradient TD method, although the connection to mirror descent and Bregman divergences were not made in this previous study, and EG does not generate sparse solutions [SST11a]. We discuss EG methods vs. p-norm methods in Section 6.

### 2.2 Sparse Learning with Mirror Descent TD

**Algorithm 2** Sparse Mirror Descent TD($\lambda$)

1: **repeat**
2:  Do action $\pi(s_t)$ and observe next state $s_{t+1}$ and reward $r_t$.
3:  Update the eligibility trace $e_t \leftarrow e_t + \lambda\gamma\phi(s_t)$
4:  Update the dual weights $\theta_t$:

$$\tilde{\theta}_{t+1} = \nabla\psi_t(w_t) + \alpha_t \left(r_t + \gamma\phi(s_{t+1})^T w_t - \phi(s_t)^T w_t\right) e_t$$

   (e.g., $\psi(w) = \frac{1}{2}\|w\|_q^2$ is the p-norm link function).
5:  Truncate weights:

$$\forall j, \quad \theta_j^{t+1} = \text{sign}(\tilde{\theta}_j^{t+1}) \max(0, |\tilde{\theta}_j^{t+1}| - \alpha_t\beta)$$

6:  $w_{t+1} = \nabla\psi_t^*(\theta_{t+1})$ (e.g., $\psi^*(\theta) = \frac{1}{2}\|\theta\|_p^2$ and $p$ and $q$ are dual norms such that $\frac{1}{p} + \frac{1}{q} = 1$).
7:  Set $t \leftarrow t + 1$.
8: **until done**.
   Return $\hat{V}^\pi \approx \Phi w_t$ as the $l_1$ penalized sparse value function associated with policy $\pi$ for MDP $M$.

Algorithm 2 describes a modification to obtain sparse value functions resulting in a sparse mirror-descent TD($\lambda$) algorithm. The main difference is that the dual weights $\theta$ are truncated according to Equation 9 to satisfy the $l_1$ penalty on the weights. Here, $\beta$ is the sparsity parameter defined in Equation 7. An analogous approach was suggested in [SST11a] for $l_1$ penalized classification and regression.

### 2.3 Composite Mirror Descent TD

Another possible mirror-descent TD algorithm uses as the distance-generating function a Mahalanobis distance derived from the subgradients generated during actual trials. We base our derivation on the composite mirror-descent approach proposed in [DHS11] for classification and regression. The composite mirror-descent solves the following optimization problem at each step:

$$w_{t+1} = \text{argmin}_{x \in X} \left(\alpha_t\langle x, \partial f_t\rangle + \alpha_t\mu(x) + D_{\psi_t}(x, w_t)\right) \quad (16)$$

Here, $\mu$ serves as a fixed regularization function, such as the $l_1$ penalty, and $\psi_t$ is the time-dependent distance generating function as in mirror descent. We now describe a dif-ferent Bregman divergence to be used as the distance generating function in this method. Given a positive definite matrix $A$, the Mahalanobis norm of a vector $x$ is defined as $\|x\|_A = \sqrt{\langle x, Ax\rangle}$. Let $g_t = \partial f(s_t)$ be the subgradient of the function being minimized at time $t$, and $G_t = \sum_t g_t g_t^T$ be the covariance matrix of outer products of the subgradients. It is computationally more efficient to use the diagonal matrix $H_t = \sqrt{\text{diag}(G_t)}$ instead of the full covariance matrix, which can be expensive to estimate. Algorithm 3 describes the adaptive subgradient mirror descent TD method.

---

**Algorithm 3** Composite Mirror Descent TD($\lambda$)

1: **repeat**
2:     Do action $\pi(s_t)$ and observe next state $s_{t+1}$ and reward $r_t$.
3:     Set TD error $\delta_t = r_t + \gamma\phi(s_{t+1})^T w_t - \phi(s_t)^T w_t$
4:     Update the eligibility trace $e_t \leftarrow e_t + \lambda\gamma\phi(s_t)$
5:     Compute TD update $\xi_t = \delta_t e_t$.
6:     Update feature covariance

$$G_t = G_{t-1} + \phi(s_t)\phi(s_t)^T$$

7:     Compute Mahalanobis matrix $H_t = \sqrt{\text{diag}(G_t)}$.
8:     Update the weights $w$:

$$w_{t+1,i} = \text{sign}(w_{t,i} - \frac{\alpha_t \xi_{t,i}}{H_{tt,i}})(|w_{t,i} - \frac{\alpha_t \xi_{t,i}}{H_{tt,i}}| - \frac{\alpha_t \beta}{H_{tt,i}})$$

9:     Set $t \leftarrow t + 1$.
10: **until done**.
    Return $\hat{V}^\pi \approx \Phi w_t$ as the $l_1$ penalized sparse value function associated with policy $\pi$ for MDP $M$.

---

## 3 Convergence Analysis

**Definition 2** [GLMH11]: $\Pi_{l_1}$ is the $l_1$-regularized projection defined as: $\Pi_{l_1} y = \Phi \alpha$ such that $\alpha = \arg\min_w \|y - \Phi w\|^2 + \beta\|w\|_1$, which is a non-expansive mapping w.r.t weighted $l_2$ norm induced by the on-policy sample distribution setting, as proven in [GLMH11]. Let the approximation error $f(y, \beta) = \|y - \Pi_{l_1} y\|^2$.

**Definition 3** (Empirical $l_1$-regularized projection): $\hat{\Pi}_{l_1}$ is the empirical $l_1$-regularized projection with a specific $l_1$ regularization solver, and satisfies the non-expansive mapping property. It can be shown using a direct derivation that $\hat{\Pi}_{l_1}\Pi T$ is a $\gamma$-contraction mapping. Any unbiased $l_1$ solver which generates intermediate sparse solution before convergence, e.g., SMIDAS solver after $t$-th iteration, comprises an empirical $l_1$-regularized projection.

**Theorem 1** The approximation error $\|V - \hat{V}\|$ of Algorithm 2 is bounded by (ignoring dependence on $\pi$ for simplicity):

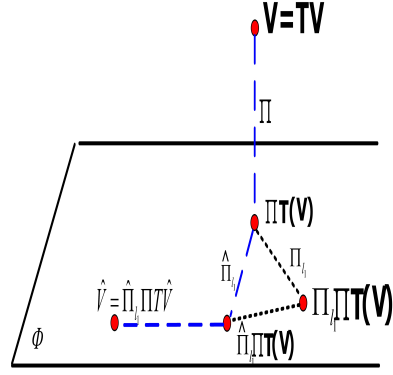

Figure 1: Error Bound and Decomposition

$$\|V - \hat{V}\| \leq \frac{1}{1-\gamma} \times$$
$$\left(\|V - \Pi V\| + f(\Pi V, \beta) + (M-1)P(0) + \|w^*\|_1^2 \frac{M}{\alpha_t N}\right) \quad (17)$$

where $\hat{V}$ is the approximated value function after $N$-th iteration, i.e., $\hat{V} = \Phi w_N$, $M = \frac{2}{2 - 4\alpha_t(p-1)e}$, $\alpha_t$ is the stepsize, $P(0) = \frac{1}{N}\sum_{i=1}^{N}\|\Pi V(s_i)\|_2^2$, $s_i$ is the state of $i$-th sample, $e = d^{\frac{p}{2}}$, $d$ is the number of features, and finally, $w^*$ is $l_1$-regularized projection of $\Pi V$ such that $\Phi w^* = \Pi_{l_1}\Pi V$.

**Proof:** In the on-policy setting, the solution given by Algorithm 2 is the fixed point of $\hat{V} = \hat{\Pi}_{l_1}\Pi T\hat{V}$ and the error decomposition is illustrated in Figure 1. The error can be bounded by the triangle inequality

$$\|V - \hat{V}\| = \|V - \Pi TV\| + \|\Pi TV - \hat{\Pi}_{l_1}\Pi TV\| + \|\hat{\Pi}_{l_1}\Pi TV - \hat{V}\| \quad (18)$$

Since $\hat{\Pi}_{l_1}\Pi T$ is a $\gamma$-contraction mapping, and $\hat{V} = \hat{\Pi}_{l_1}\Pi T\hat{V}$, we have

$$\|\hat{\Pi}_{l_1}\Pi TV - \hat{V}\| = \|\hat{\Pi}_{l_1}\Pi TV - \hat{\Pi}_{l_1}\Pi T\hat{V}\| \leq \gamma\|V - \hat{V}\| \quad (19)$$

So we have

$$(1-\gamma)\|V - \hat{V}\| \leq \|V - \Pi TV\| + \|\Pi TV - \hat{\Pi}_{l_1}\Pi TV\|$$

$\|V - \Pi TV\|$ depends on the expressiveness of the basis $\Phi$, where if $V$ lies in $span(\Phi)$, this error term is zero. $\|\Pi TV - \Pi_{l_1}\hat{\Pi}TV\|$ is further bounded by the triangle inequality

$$\|\Pi TV - \hat{\Pi}_{l_1}\Pi TV\| \leq$$
$$\|\Pi TV - \Pi_{l_1}\Pi TV\| + \|\Pi_{l_1}\Pi TV - \hat{\Pi}_{l_1}\Pi TV\|$$

where $\|\Pi TV - \Pi_{l_1}\Pi TV\|$ is controlled by the sparsity parameter $\beta$, i.e., $f(\Pi TV, \beta) = \|\Pi TV - \Pi_{l_1}\Pi TV\|$, where $\varepsilon = \|\hat{\Pi}_{l_1}\Pi TV - \Pi_{l_1}\Pi TV\|$ is the approximation error depending on the quality of the $l_1$ solver employed. In Algorithm 2, the $l_1$ solver is related to the SMIDAS $l_1$ regularized mirror-descent method for regression and classification [SST11a]. Note that for a squared loss function

$L(\langle w, x_i \rangle, y_i) = ||\langle w, x_i \rangle - y_i||_2^2$, we have $|L'|^2 \leq 4L$. Employing the result of Theorem 3 in [SST11a], after the $N$-th iteration, the $l_1$ approximation error is bounded by

$$\varepsilon \leq (M-1)P(0) + ||w^*||_1^2 \frac{M}{\alpha_t N}, M = \frac{2}{2 - 4\alpha_t(p-1)e}$$

By rearranging the terms and applying $V = TV$, Equation (17) can be deduced.

## 4 Experimental Results: Discrete MDPs

Figure 2 shows that mirror-descent TD converges more quickly with far smaller Bellman errors than LARS-TD [KN09] on a discrete "two-room" MDP [MM07]. The basis matrix $\Phi$ was automatically generated as 50 proto-value functions by diagonalizing the graph Laplacian of the discrete state space connectivity graph[MM07]. The figure also shows that Algorithm 2 (sparse mirror-descent TD) scales more gracefully than LARS-TD. Note LARS-TD is unstable for $\gamma = 0.9$. It should be noted that the computation cost of LARS-TD is $O(Ndm^3)$, whereas that for Algorithm 2 is $O(Nd)$, where $N$ is the number of samples, $d$ is the number of basis functions, and $m$ is the number of active basis functions. If $p$ is linear or sublinear w.r.t $d$, Algorithm 2 has a significant advantage over LARS-TD.

Figure 3 shows the result of another experiment conducted to test the noise immunity of Algorithm 2 using a discrete $10 \times 10$ grid world domain with the goal set at the upper left hand corner. For this problem, 50 proto-value basis functions were automatically generated, and 450 random Gaussian mean 0 noise features were added. The sparse mirror descent TD algorithm was able to generate a very good approximation to the optimal value function despite the large number of irrelevant noisy features, and took a fraction of the time required by LARS-TD.

Figure 4 compares the performance of mirror-descent Q-learning with a fixed p-norm link function vs. a decaying p-norm link function for a $10 \times 10$ discrete grid world domain with the goal state in the upper left-hand corner. Initially, $p = O(\log d)$ where $d$ is the number of features, and subsequently $p$ is decayed to a minimum of $p = 2$. Varying $p$-norm interpolates between additive and multiplicative updates. Different values of $p$ yield an interpolation between the truncated gradient method [LLZ09] and SMIDAS [SsT09].

Figure 5 illustrates the performance of Algorithm 3 on the two-room discrete grid world navigation task.

## 5 Experimental Results: Continuous MDPs

Figure 6 compares the performance of Q-learning vs. mirror-descent Q-learning for the mountain car task, which converges more quickly to a better solution with much lower variance. Figure 7 shows that mirror-descent Q-learning with learned diffusion wavelet bases converges quickly on the 4-dimensional Acrobot task. We found in our experiments that LARS-TD did not converge within 20 episodes (its curve, not shown in Figure 6, would be flat on the vertical axis at 1000 steps). Finally, we tested the mirror-descent approach on a more complex 8-dimensional continuous MDP. The triple-link inverted pendulum [SW01] is a highly nonlinear time-variant underactuated system, which is a standard benchmark testbed in the control community. We base our simulation using the system parameters described in [SW01], except that the action space is discretized because the algorithms described here are restricted to policies with discrete actions. There are three actions, namely $\{0, 5\text{Newton}, -5\text{Newton}\}$. The state space is 8-dimensional, consisting of the angles made to the horizontal of the three links in the arm as well as their angular velocities, the position and velocity of the cart used to balance the pendulum. The goal is to learn a policy that can balance the system with the minimum num-ber of episodes. A run is successful if it balances the inverted pendulum for the specified number of steps within $300$ episodes, resulting in a reward of $0$. Otherwise, this run is considered as a failure and yields a negative reward $-1$. The first action is chosen randomly to push the pendulum away from initial state. Two experiments were conducted on the triple-link pendulum domain with $20$ runs for each experiment. As Table 1 shows, Mirror Descent Q-learning is able to learn the policy with fewer episodes and usually with reduced variance compared with regular Q-learning.

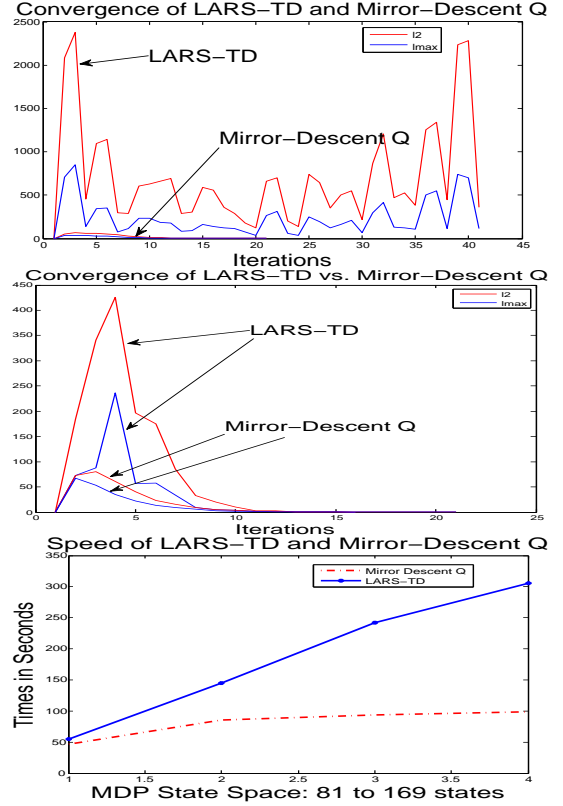

Figure 2: Mirror-descent Q-learning converges significantly faster than LARS-TD on a "two-room" grid world MDP for $\gamma = 0.9$ (top left) and $\gamma = 0.8$ (top right). The y-axis measures the $l_2$ (red curve) and $l_\infty$ (blue curve) norm difference between successive weights during policy iteration. Bottom: running times for LARS-TD (blue solid) and mirror-descent Q (red dashed). Regularization $\beta = 0.01$.

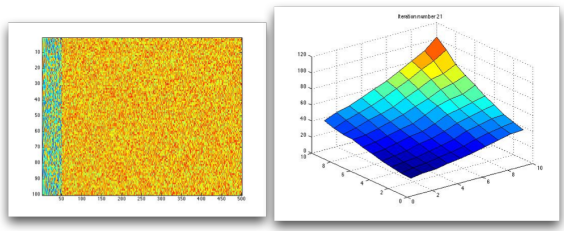

Figure 3: Sensitivity of sparse mirror-descent TD to noisy features in a grid-world domain. Left: basis matrix with the first 50 columns representing proto-value function bases and the remainder 450 bases representing mean-0 Gaussian noise. Right: Approximated value function using sparse mirror-descent TD.

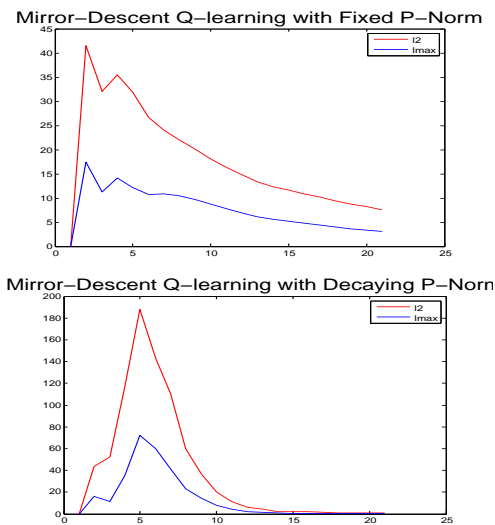

Figure 4: Left: convergence of mirror-descent Q-learning with a fixed p-norm link function. Right: decaying p-norm link function.

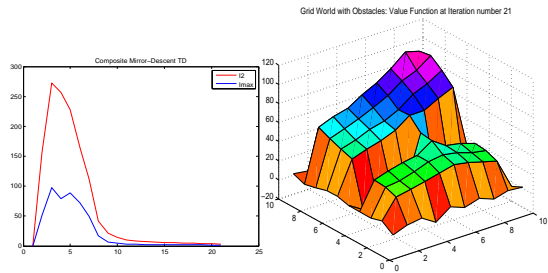

Figure 5: Left: Convergence of composite mirror-descent Q-learning on two-room gridworld domain. Right: Approximated value function, using $50$ proto-value function bases.

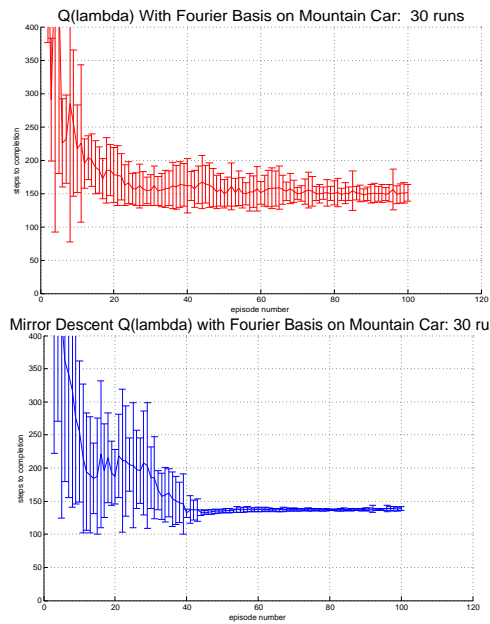

Figure 6: Top: Q-learning; Bottom: mirror-descent Q-learning with p-norm link function, both with 25 fixed Fourier bases [KOT08] for the mountain car task.

The experiment settings are Experiment 1: Zero initial state and the system receives a reward $1$ if it is able to balance 10,000 steps. Experiment 2: Zero initial state and the system receives a reward $1$ if it is able to balance $100,000$ steps. Table 1 shows the comparison result between regular Q-learning and Mirror Descent Q-learning.

## 6 Comparison of Link Functions

The two most widely used link functions in mirror descent are the $p$-norm link function [BT03] and the relative entropy function for exponentiated gradient (EG) [KW95]. Both of these link functions offer a multiplicative update rule compared with regular additive gradient methods. The differences between these two are discussed here. Firstly, the loss function for EG is the relative entropy whereas that of the $p$-norm link function is the square $l_2$-norm function. Second and more importantly, EG does not produce sparse solutions since it must maintain the weights away from zero, or else its potential (the relative entropy) becomes unbounded at the boundary.

Another advantage of $p$-norm link functions over EG is that the $p$-norm link function offers a flexible interpolation between additive and multiplicative gradient updates. It has been shown that when the features are dense and the optimal coefficients $\theta^*$ are sparse, EG converges faster than the regular additive gradient methods [KW95]. However, according to our experience, a significant drawback of EG is the overflow of the coefficients due to the exponential operator. To prevent overflow, the most commonly used technique is rescaling: the weights are re-normalized to sum to a constant. However, it seems that this approach does not always work. It has been pointed out [PS95] that in the EG-Sarsa algorithm, rescaling can fail, and replacing eligible traces instead of regular additive eligible traces is used to prevent overflow. EG-Sarsa usually poses restrictions on the basis as well. Thanks to the flexible interpolation capability between multiplicative and additive gradient updates, the $p$-norm link function is more robust and applicable to various basis functions, such as polynomial, radial basis function (RBF), Fourier basis [KOT08], proto-value functions (PVFs), etc.

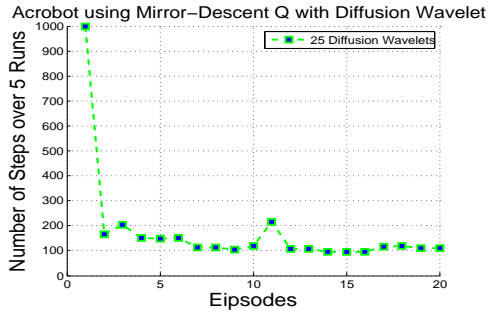

Figure 7: Mirror-descent Q-learning on the Acrobot task using automatically generated diffusion wavelet bases averaged over 5 trials.

Table 1: Results on Triple-Link Inverted Pendulum Task.

| # of Episodes\Experiment | 1 | 2 |
|---|---|---|
| Q-learning | $6.1 \pm 5.67$ | $15.4 \pm 11.33$ |
| Mirror Descent Q-learning | $5.7 \pm 9.70$ | $11.8 \pm 6.86$ |

## 7 Summary and Future Work

We proposed a novel framework for reinforcement learning using mirror-descent online convex optimization. Mirror Descent Q-learning demonstrates the following advantage over regular Q learning: faster convergence rate and reduced variance due to larger stepsizes with theoretical convergence guarantees [NJLS09]. Compared with existing sparse reinforcement learning algorithms such as LARS-TD, Algorithm 2 has lower sample complexity and lower computation cost, advantages accrued from the first-order mirror descent framework combined with proximal mapping [SST11a]. There are many promising future research topics along this direction. We are currently exploring a mirror-descent fast-gradient RL method, which is both convergent off-policy and quicker than fast gradient TD methods such as GTD and TDC [SMP$^+$09]. To scale to large MDPs, we are investigating hierarchical mirror-descent RL methods, in particular extending SMDP Q-learning. We are also undertaking a more detailed theoretical analysis of the mirror-descent RL framework, building on existing analysis of mirror-descent methods [DHS11, SST11a]. Two types of theoretical investigations are being explored: regret bounds of mirror-descent TD methods, extending previous results [SW94] and convergence analysis combining robust stochastic approximation [NJLS09] and RL theory [BT96, Bor08].

## Acknowledgments

This material is based upon work supported by the Air Force Office of Scientific Research (AFOSR) under grant FA9550-10-1-0383, and the National Science Foundation under Grant Nos. NSF CCF-1025120, IIS-0534999, and IIS-0803288. Any opinions, findings, and conclusions or recommendations expressed in this material are those of the authors and do not necessarily reflect the views of the AFOSR or the NSF.